# Automated detection of Zika and dengue in *Aedes aegypti* using neural spiking analysis

Danial Sharifrazi[1], Nouman Javed[1,2], Roohallah Alizadehsani[1], Prasad N. Paradkar[2], U. Rajendra Acharya[3,4], and Asim Bhatti[1]

[1]*Institute for Intelligent Systems Research and Innovations (IISRI), Deakin University, Geelong, Australia*
[2]*CSIRO Health and Biosecurity, Australian Animal Health Laboratory, Geelong, Australia*
[3]*School of Mathematics, Physics and Computing, University of Southern Queensland, Springfield, Australia*
[4]*Center for Health Research, University of Southern Queensland, Springfield, Australia*

**ABSTRACT**

Mosquito-borne diseases present considerable risks to the health of both animals and humans. *Aedes aegypti* mosquitoes are the primary vectors for numerous medically important viruses such as dengue, Zika, yellow fever, and chikungunya. To characterize this mosquito neural activity, it is essential to classify the generated electrical spikes. However, no open-source neural spike classification method is currently available for mosquitoes. Our work presented in this paper provides an innovative artificial intelligence-based method to classify the neural spikes in uninfected, dengue-infected, and Zika-infected mosquitoes. Aiming for outstanding performance, the method employs a fusion of normalization, feature importance, and dimension reduction for the preprocessing and combines convolutional neural network and extra gradient boosting (XGBoost) for classification. The method uses the electrical spiking activity data of mosquito neurons recorded by microelectrode array technology. We used data from 0, 1, 2, 3, and 7 days post-infection, containing over 15 million samples, to analyze the method's performance. The performance of the proposed method was evaluated using accuracy, precision, recall, and the F1 scores. The results obtained from the method highlight its remarkable performance in differentiating infected vs uninfected mosquito samples, achieving an average of 98.1%. The performance was also compared with 6 other machine learning algorithms to further assess the method's capability. The method outperformed all other machine learning algorithms' performance. Overall, this research serves as an efficient method to classify the neural spikes of *Aedes aegypti* mosquitoes and can assist in unraveling the complex interactions between pathogens and mosquitoes.

Keywords: *Aedes aegypti*, Dengue, Zika, Classification, Machine Learning, Deep Learning

## 1. INTRODUCTION

Arbovirus infections spread by mosquitoes significantly influence human health worldwide [1]. The *Flaviviridae* family of viruses is notable for the high prevalence of diverse arboviruses capable of infecting humans and causing illnesses, thus emphasizing their substantial clinical significance [2]. Dengue, West Nile, Japanese encephalitis, and more recently Zika [3] are the mosquito-borne viruses that cause the most severe morbidity and mortality.

Studying how mosquitoes interact with pathogens is crucial for understanding the transmission and epidemiology of mosquito-borne diseases. Viruses can affect mosquito behavioral traits, such as locomotion, fecundity, feeding, and fertility [4]. Additionally, infection with mosquito-borne viruses can also affect the nervous systems of humans and mosquitos (Gaburro 2018). Previous studies have shown that the Zika virus (ZIKV) which caused the 2014–2015 outbreaks in the Pacific and the Americas leads to neuropathology, including Guillain–Barre syndrome in French Polynesia [5] and the Americas [6], a congenital syndrome in newborns with microcephaly [7] and ocular abnormalities [8, 9], in South America [7], and recent records of encephalitis [10, 11], and myelitis in adults [12]. Numerous studies have been conducted on the neurovirulence of the ZIKV using various models, including mouse models [13, 14], human pluripotent stem cell (hPSC)-derived neural progenitor cells and organoids [15, 16], and strains of the virus from Brazil, Asia, and Africa [8, 17].

The West Nile virus can infect human neuronal cells, causing encephalitis [18], and also affects its mosquito vector's nervous system [19]. Similarly, although less neurotropic in humans than the West Nile virus, the dengue virus (DENV) displays neurotropic tendencies in mosquitoes [20] and can alter mosquito behavior [21]. ZIKV has considerable neurotropism in humans and other animal models, but the nervous system infection in the vector stage of the viral lifecycle has not yet received enough attention.

Mammal and insect neurons are conserved at many cellular levels [22], as shown by several *Drosophila* models for human neurological diseases [23]. Despite the apparent distinctions (placenta, blood-brain barrier, etc.), the two virus hosts may have similar viral infection mechanisms. Therefore, studying the neural activity of natural vector hosts is important to understanding the molecular mechanisms that could underlie disease mechanisms. The classification of neural spikes is pivotal for gaining insights into neural activity. Though different tools have been developed to monitor the behaviors of mosquitoes [24, 25], very few tools have been developed to understand mosquitoes' neural activity. Gaburro *et al.* used NeuroSigX, a custom-made tool, to classify the neural spikes of mosquitoes [26]. To the authors' best knowledge, no open-source neural spike classification method is currently available for mosquitoes.

In recent years, significant contributions have been made to neural spike classification by utilizing different machine-learning techniques. Zhang *et al.* [27] applied K-mean clustering on the synthetic neural and empirical datasets, showing that MEA recording is the preferred choice of all spike recording methods. Zubair *et al.* [28] worked on Principal Component Analysis (PCA) combined with Catboost and random forest classifiers. Ramesh *et al.* [29] presented a new method based on a multi-layer, multi-spiking neural network and mini-batch Stochastic Gradient Descent for electroencephalographic (EEG) classification. Shen *et al.* [30] used tunable-Q wavelet transform and Convolutional Neural Network (CNN) for signal classification. Jeon et al. [31] also implemented their classifier based on CNN. Joshi et al. [32] introduced a direct thresholding-based approach. Zhao *et al.* [33] showed that combining Artificial Neural Networks (ANNs) and Spiking

Neural Networks (SNNs) as a hybrid classifier for neural signal analyses can be a good idea. Shrestha et al. [34] mentioned in their paper that Spiking Neural Networks (SNNs) are promising research models for signal classification. Yan et al. [35] presented a classifier based on SNN and used a transfer learning technique to classify signal data efficiently.

Given the significance of monitoring neural activity in mosquitoes and the absence of open-source methods, we introduce a novel machine learning-based neural spike classification approach in this study. The method combines normalization, feature importance, and dimension reduction for effective preprocessing. Moreover, the classification is performed by combining two machine learning methods: Convolutional Neural Network (CNN) and XGBoost. The method was tested using the neural spikes data of uninfected, dengue-infected, and Zika-infected *Aedes aegypti* (*A. aegypti*) mosquitoes. The proposed method will also be compared with eight other well-known machine learning algorithms using metrics involving accuracy, precision, recall, F1 score, and training time to analyze the method's capability further.

## 2. DATASET USED

**Viruses strain and preparation**
We employed a human isolate of the Zika virus (ZIKV) originating from Cambodia in 2010 (GenBank KU955593). The viral isolate was subjected to serial passages in the C6-36 cell line and harvested 72 hours post-infection. For dengue virus, dengue virus serotype 2 ET300 (DENV2), isolated from an Australian soldier returning from East Timor was used. The isolate underwent seven passages in the C6-36 cell line.

**Primary neurons preparation**
*A. aegypti* adult female mosquitoes were cold-killed three to four days after emergence, and the brains were then dissected and separated in L15 medium supplemented with penicillin-streptomycin (50 units/ml), fungizone, 10% tryptase phosphate broth, and 10% fetal calf serum. A 2 ml tube with a clean medium and fifteen to twenty brains were combined and placed inside for mechanical dissociation by repeated trituration. The cells were then resuspended in 50 µl of clean medium after being centrifuged twice (5 x g for 3 minutes).

**Cultures of primary neurons and infections**
As previously mentioned, primary neuron cultures were established for confocal microscopy tests and $TCID_{50}$. Coverslips were precoated for 30 minutes at room temperature with 100 µl of polyethyleneimine (PEI, Sigma, 0.05% in Borate-buffered solution), washed three times with tissue culture treated water (TC-water), and allowed to dry in the Biosafety Cabinet class II (BSCII). In a 24-well plate, cells were plated using 25 µl of a stock solution containing $3 \times 10^6$ cells per ml. Before adding 1 ml of finished L15 medium (10% FCS), plates were incubated at 28 °C for 30 minutes to allow cells to adhere.

The technique was adapted from Hales et al.'s [36] for primary neuron culturing using microelectrode arrays (MEA). Before the previously mentioned cell seeding, the microelectrode array was precoated using 100 µl of PEI for coating, which was then followed by three TC-water rinses. Finally, just before plating, laminin (0.02 mg/ml, Sigma L-2020) was added for 20 minutes at 37 °C and 5% $CO_2$. In the middle of the MEAs, cells were plated at a density of $7.5 \times 10^4$ per unit (25 µl volume). 1 ml of cell medium was introduced after cells had been allowed to attach to the bottom of the MEA well for 30 minutes in the 28 °C incubator. Every three to four days, with the day before recording as an exception, half of the culture medium was replaced. At 7 days of in vitro (div) culture, following network maturity, all primary neuron cultures were infected with the Multiplicity of Infection (MOI) of 0.2. Infections of 24-well plate cultures were treated by removing the whole medium and replacing it with 1 ml of the virus. Cells were infected with 0.2 MOI virus in 300 µl for an hour before new media was added for infection on MEAs.

**Utilizing a microelectrode array to acquire data**

*Recording material:* Microelectrode arrays (MEAs) from Multi-Channel Systems, Reutlingen, Germany (60MEA 200/30iR-Ti-gr), were employed for the electrophysiological recording of neuronal networks. The MEA comprises 59 planar round electrodes of TiN/SiN, featuring diameters ranging from 10 to 30 µm and center-to-center inter-electrode intervals of 100 to 200 µm, arranged in a square grid format without corners. One recording electrode was replaced by a single, bigger electrode that acted as the reference ground electrode. Online cellular activity was captured using the MEA60inv System (Multi Channel Systems, Reutlingen, Germany). The action potential was captured at each electrode using a 10 kHz sampling rate.

*Recording technique:* For the safe containment of viruses, all operations and recordings were done in a Biosafety Cabinet class II (BSCII). Each recording session involved capturing 30 minutes of spontaneous activity at 0, 2, 3, and 7 dpi. A 15-minute recording of post-stimulus activity was conducted at 8 dpi for gabazine stimulation assays. Each recording session began 5-10 minutes after the MEA was positioned on the amplifier, aiming to prevent neuronal responses to mechanical stress from movement. This interval also allowed for adaptation to the BSCII conditions and the temperature maintained by the adapter instead of the incubator. The temperature of the MEA was kept at 28 ºC throughout the recording to align with the optimal temperature for the viability of the neuronal culture. After each recording session, half of the media was replaced.

The raw continuous voltage recordings underwent filtration to eliminate field potential traces below 200 Hz, which were produced by the collectively charged network. A 2nd order Butterworth filter with a 200 Hz cutoff frequency made up the high pass filter. Spike detection was achieved by setting the threshold at eight times the standard deviation of the high-pass filtered signal [37], measured at 22 µV using the MC_Rack software, within a 500 ms window.

*Recording time points:* After being plated on the MEA, the dissociated neurons from mosquitoes were given seven days to nourish and develop. *A. aegypti* primary neuron networks were infected with ZIKV or DENV2 (used as a control) at an MOI of 0.2 at 7 days in vitro (div).

Six electrophysiological recordings of spontaneous activity were made for each experimental test. The first was completed at 7 div before infection and used as a benchmark for further research. Records were sampled two, three, and seven days after infection (dpi). At 8 dpi, a GABAA antagonist called gabazine [30] (Tocris SR 95531 hydrobromide [2- (3′-carboxy-2′-propyl)-3-amino-6-p-methoxyphenylpyridazinium bromide]) was used to stimulate the neurons.

## Programming and computing system

Python programming language and Keras package with TensorFlow backend were used to develop the proposed method. The NVIDIA GeForce RTX 4080 GPU was hired to train and test the CNN and the Intel Core i9 2.20 GHz processor (13[th] Generation) was utilized to train and test machine learning methods.

## Dataset

The dataset employed in this research comprises a total of 15,728,580 samples, encompassing Control, DENV2, and ZIKV classes across various time points, 0, 1, 2, 3, and 7 days post-infection. Notably, the exact utilized data for any given class on any day post-infection was 1,048,572 samples. This highlights the depth of the dataset used. Each of these samples corresponds to the signals recorded in the unit of time. The tool (MEA) used to record signals has 60 different channels, and the data extracted from each channel is equivalent to one feature in the dataset. Some samples of the dataset can be visualized in Figure 1.

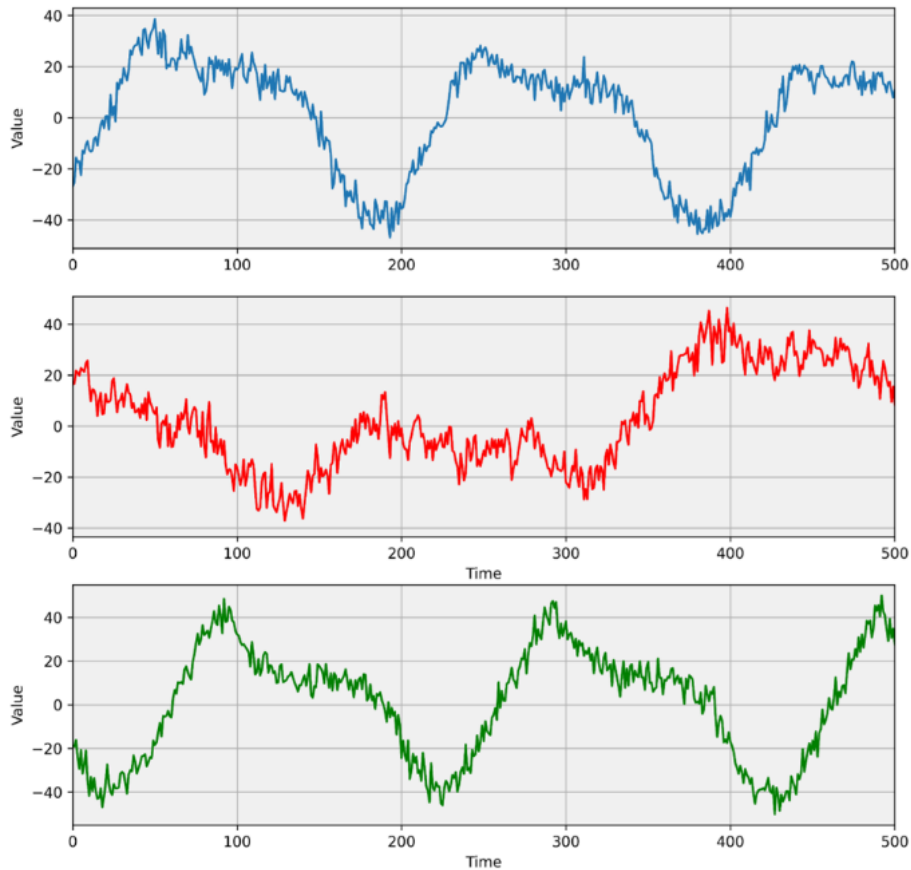

Figure 1. Dataset visualization. The blue color indicates the control class, while the red and green depict the DENV and ZIKV, respectively.

## K-fold cross-validation

To ensure the consistency of the model, we employed the k-fold cross-validation method with k set to 10. This means the dataset was divided into 10 subsets. The distribution within these subsets comprised 80% for training, 10% for validation, and another 10% for testing. The process involved multiple iterations, with each subset taking turns as the test set. After each round, the model's performance metrics were recorded, and in the end, the averages of these metrics were computed. This approach provides a robust assessment of the model, enhancing our confidence in its ability to perform well with new, unseen data, as it has been tested on various sections of the dataset.

**Evaluation metrics**

The performance of the proposed method was evaluated using widely recognized evaluation metrics for the classification tasks. The metrics encompass accuracy, precision, recall, and the F1 score, each providing unique insights into the model's performance. The formulas for calculating these metrics are detailed below [38].

$$\text{Accuracy} = \frac{(TP+TN)}{(TP+TN+FP+FN)} \qquad (1)$$

$$\text{Precision} = \frac{TP}{(TP+FP)} \qquad (2)$$

$$\text{Recall} = \frac{TP}{(TP+FN)} \qquad (3)$$

$$\text{F1\_Score} = \frac{2 \times (\text{Precision} \times \text{Recall})}{(\text{Precision}+\text{Recall})} \qquad (4)$$

Where TP represents true positives (correctly predicted positive classes), TN denotes true negatives (correctly predicted negative classes), FP signifies false positives (incorrectly predicted positive classes), and FN stands for false negatives (incorrectly predicted negative classes)

Furthermore, the performance analysis was expanded by plotting precision-recall curves, visually representing the model's precision and recall trade-offs.

## 3. METHOD

The method employs the input dataset encompassing Uninfected control, DENV2, and ZIKV classes, which is then divided into 10 subsets using k-fold cross-validation. Subsequently, the data undergoes two stages: the preprocessing stage and the classification stage. The preprocessing stage comprises normalization, feature importance, and dimension reduction, while the classification stage combines the CNN and XGBoost (Figure 2). The preprocessing and classification stages are discussed in detail below.

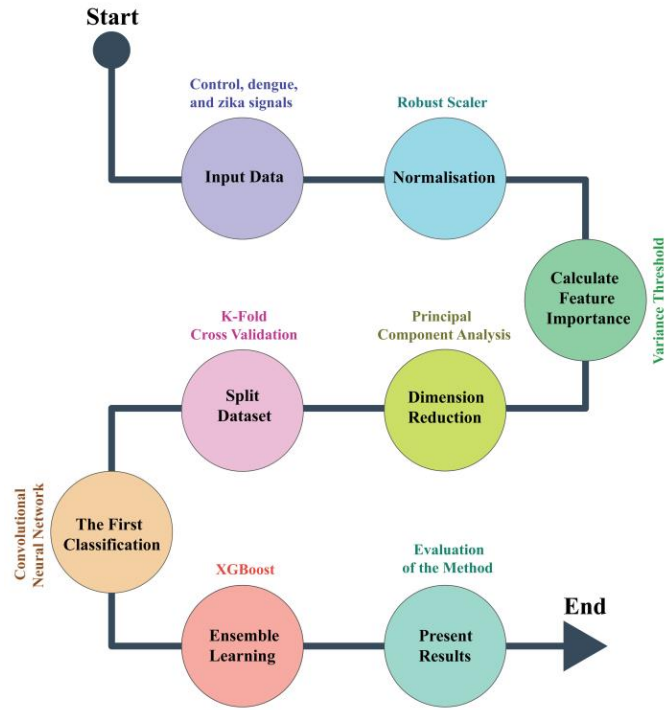

Figure 2: Proposed methodology.

**Preprocessing stage:**

*Normalization*

The preprocessing stage starts from normalization, a crucial step to ensure that features with varied scales contribute equitably to the model's learning process, preventing the domination of certain features. Considering outliers in our dataset, our methodology employs the Robust Scaler method for the normalization, specifically chosen for its reduced sensitivity to outliers. The Robust Scaler's use of the interquartile range for scaling ensures a more robust treatment of extreme values. The Robust Scaler method available in the Scikit-Learn package in Python was used for this purpose.

*Feature importance*

Next, the feature importance was calculated. Feature importance evaluates the significance of input features for predicting outcomes. It helps identify key factors influencing the model's predictions, aiding in effective feature selection. The Variance Threshold method was used to calculate the feature importance in the proposed method. The results obtained from the feature importance calculations are shown in Figure 3-a.

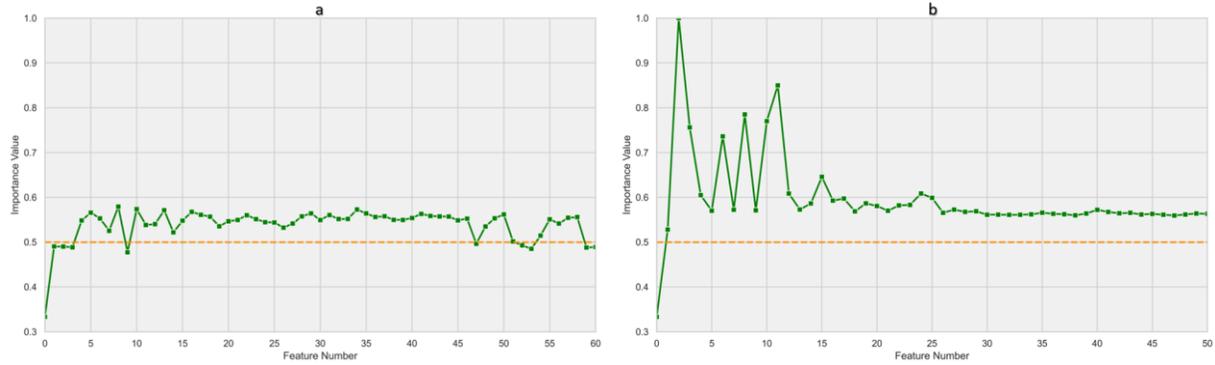

Figure 3: Feature importance values: (a) Obtained based on initial calculations. And (b) Obtained after performing PCA.

Following the feature importance calculations, a threshold of 0.5 was set to assess which features surpassed this criterion. Of the initial 61 features, 51 exhibited feature importance values exceeding the 0.5 threshold, while the remaining 10 features fell below (Figure 3-a). Based on these results, one approach could have been excluding the low-importance features and proceeding with the high-importance ones. However, an alternative strategy, dimension reduction, was adopted.

*Dimension reduction*

The Principal Component Analysis (PCA) was applied, allowing the retention of all features while reducing dimensionality. This choice aimed to capture essential information even from low-importance features, providing a more comprehensive representation of the dataset. The PCA method reduced the dimensions of the dataset from 61 dimensions to 51 dimensions. Here, dimensions represent the features. After reducing the dimensions of the dataset from 61 to 51, feature importance calculations were reperformed. This time, out of 51 new features, 50 have shown feature importance values higher than the 0.5 threshold (Figure 3-b). The only feature that has shown a feature importance value lower than 0.5 was related to time.

Further, to confirm the effectiveness of the dimension reduction method, a tuned CNN model, which will be discussed in the next section, was applied to both the 61 features data and the dimensionally reduced (51 dimensions) data. The dimensionally reduced data has shown 97% accuracy (Figure 4-b) compared to the 93% accuracy (Figure 4-a) of the 61 features' data. Hence, the dimension reduction method helped to perform effective preprocessing without affecting the number of features.

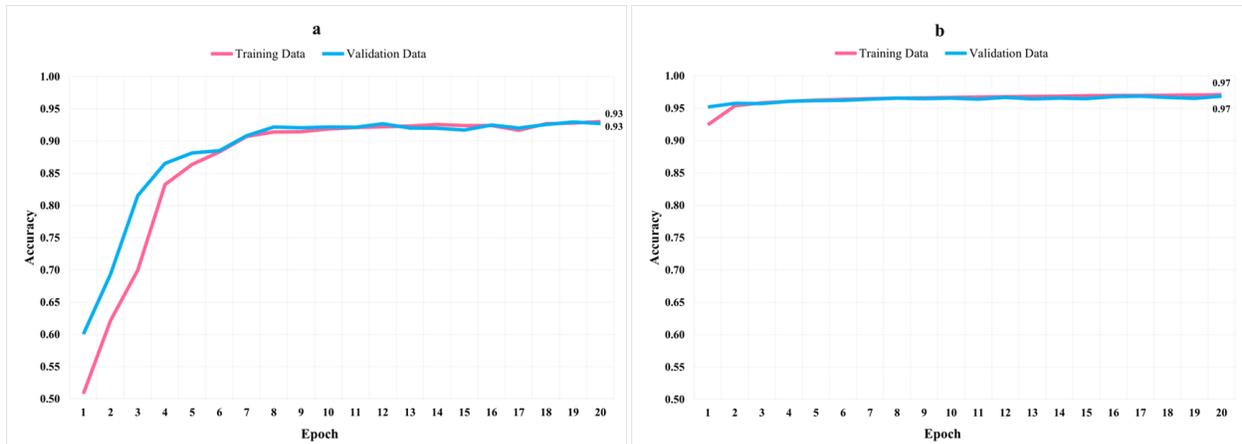

Figure 4: Training process of the convolutional neural network before and after dimension reduction. (a) Convolutional neural network accuracy before dimension reduction. (b) Convolutional neural network accuracy after dimension reduction.

## Classification stage:

### *Convolutional Neural Network (CNN)*

This section discusses the CNN of the proposed method. The initial phase involved the selection of the most suitable optimizer for the CNN concerning the dataset. A comprehensive assessment was conducted, testing the performance of 7 distinct optimizers, namely Adam, Adadelta, Adagrad, Adamax, Nadam, SGD (Stochastic Gradient Descent), and RMSprop (Root Mean Square Propagation). The assessment revealed that Adam Optimizer outperformed its counterparts, showcasing superior accuracy (96.88%) (Table 1).

Table 1. Obtained evaluation metrics by different optimizers.

| Optimizer | Accuracy (%) | Precision (%) | Recall (%) | F1 Score (%) | Training Time (m:s) |
|---|---|---|---|---|---|
| SGD | 95.98 | 95.98 | 95.98 | 95.98 | 02:02 |
| RMSprop | 95. 44 | 95.60 | 95.44 | 95.44 | 04:29 |
| Nadam | 92.11 | 92.15 | 92.11 | 92.12 | 06:44 |
| Adadelta | 72.14 | 71.66 | 72.12 | 71.77 | 02:12 |
| Adamax | 92.28 | 92.29 | 92.28 | 92.28 | 02:24 |
| Adagrad | 92.17 | 92.22 | 92.17 | 92.19 | 02:05 |
| **Adam** | **96.88** | **96.88** | **96.88** | **96.88** | **02:27** |

After selecting the right optimizer, the subsequent step focused on determining the optimal number of convolutional and fully connected (dense) layers. The convolutional neural network we used was initially configured to 1 input layer, 1 output layer, 3 convolutional layers, and 3 dense layers. First, we started finding the optimal number of convolutional layers. We assessed the metrics by varying the number of convolutional layers from 1 to 7. The outcome showed the best option was CNN with 3 convolutional layers (Table 2). So, 3 convolutional layers were selected for the CNN of the proposed method.

Table 2. Obtained evaluation metrics using different convolutional layers.

| Number of Convolutional Layers | Accuracy (%) | Precision (%) | Recall (%) | F1 Score (%) |
|---|---|---|---|---|
| 1 | 96.65 | 96.66 | 96.65 | 96.65 |
| 2 | 96.73 | 96.74 | 96.73 | 96.73 |
| **3** | **96.88** | **96.88** | **96.88** | **96.88** |
| 4 | 96.33 | 96.35 | 96.33 | 96.34 |
| 5 | 94.83 | 94.84 | 94.83 | 94.83 |
| 6 | 94.76 | 94.77 | 94.76 | 94.75 |
| 7 | 94.91 | 94.93 | 94.91 | 94.92 |

The next step was selecting the optimal number of dense layers. Similar to the convolutional layers selection method, the number of dense layers varied from 1 to 7, and accuracy was observed. The number of dense layers had no significant effect on the accuracy. However, the best one was 3 layers. Regarding CNN with 3 and 4 layers, undoubtedly CNN with fewer layers causes less training and testing time. Therefore, we decided to use 3 dense layers to avoid the increase in the training time of CNN. Table 3 illustrates the obtained evaluation metrics for 20 epochs for different numbers of dense layers.

Table 3. Obtained evaluation metrics using different dense layers.

| Number of Dense Layers | Accuracy (%) | Precision (%) | Recall (%) | F1 Score (%) |
|---|---|---|---|---|
| 1 | 96.75 | 96.75 | 96.75 | 96.75 |
| 2 | 96.67 | 96.69 | 96.67 | 96.67 |
| **3** | **96.88** | **96.88** | **96.88** | **96.88** |
| 4 | 96.88 | 96.88 | 96.88 | 96.88 |
| 5 | 96.76 | 96.76 | 96.76 | 96.76 |
| 6 | 96.61 | 96.63 | 96.61 | 96.61 |
| 7 | 96.73 | 96.73 | 96.73 | 96.73 |

The final configuration of the CNN of the proposed method was 1 input layer, 1 output layer, 3 convolutional layers, and 3 dense layers. Adam was used as an optimizer. Other hyperparameters used in the CNN of the proposed method are shown in Table 4.

Table 4. Hyperparameters used by the CNN architecture.

| **Hyper-parameters** | **Values** |
|---|---|
| Input dimension | 51*1 |
| Number of neurons of the first layer | 51*1 |
| Number of filters in convolutional layers | 64, 128, 256 |
| Strides | 2 |
| Number of neurons of fully connected layers | 256,128,64 |
| Layers activation function | ReLu |
| Optimizer function | Adam |
| Learning rate | 0.001 |
| Loss function | Categorical Cross-entropy |
| Number of epochs | 20 |
| Batch size | 1024 |

*Extreme Gradient Boosting (XGBoost)*

XGBoost is a predictive algorithm that uses decision trees and gradient boosting for making predictions. The accuracy achieved after preprocessing and fine-tuning the CNN model was notably high (98%). However, we decided to integrate the XGBoost into the CNN model's output. This choice stemmed from prior research suggesting enhanced performance in human brain signals and transformer faults classification by integrating the CNN model and XGBoost [39, 40]. The CNN model's output was subsequently fed into the XGBoost, resulting in a substantial accuracy boost to an impressive 98.1%. This collaborative approach demonstrates the effectiveness of combining the CNN model with XGBoost to achieve superior performance in the classification tasks.

In general, the proposed method includes different steps in preprocessing and classification tasks which are executed consecutively. The proposed algorithm was implemented by Python programming language 3.9. However, for a better understanding of how the program runs, a pseudocode of the algorithm is presented in Table 5 to explain the algorithm in detail.

Table 5: Pseudocode of the proposed method

| Detection of DENV2 and ZIKV infection using the proposed algorithm |
|---|
| **Input:** Feeding the algorithm with the proposed dataset including approximately 15 million samples and 61 features (dimensions)<br>**Output:** Recognized class (Control/DENV2/ZIKV) for each sample<br>　1. **Begin**<br>　2. Feeding the proposed algorithm with 0 dpi data<br>　3. Applying the normalization method using Robust Scaler<br>　4. Calculating feature importance using Variance Threshold<br>　5. Obtaining the number of the most important features (51 features)<br>　6. Reducing the data dimensions from 61 to 51 using PCA<br>　7. Splitting the dataset using a 10-fold cross-validation technique<br>　8. For naïve in range (fold=1:10):<br>　　　a. Training the CNN model<br>　　　b. Passing the dataset through the CNN<br>　　　c. Obtaining new features from CNN's output layer<br>　　　d. Training the XGBoost model with new features<br>　　　e. Testing the XGBoost model<br>　　　f. Obtaining outputs from the XGBoost<br>　9. Repeating the algorithm from step 2 to step 8 for 1, 2, 3, and 7 dpi data<br>　10. **End** |

# 4. RESULTS

As previously briefly mentioned, the performance following preprocessing and integrating the XGBoost into the CNN was high. Detailed results are presented here. Figure 5 displays the confusion matrix, where dark green boxes signify accurate classification for the Control, DENV2, and ZIKV classes, and pale green boxes represent the instances of incorrect predictions. It can be seen that the number of correct predictions is very high for all classes for all days post-infection.

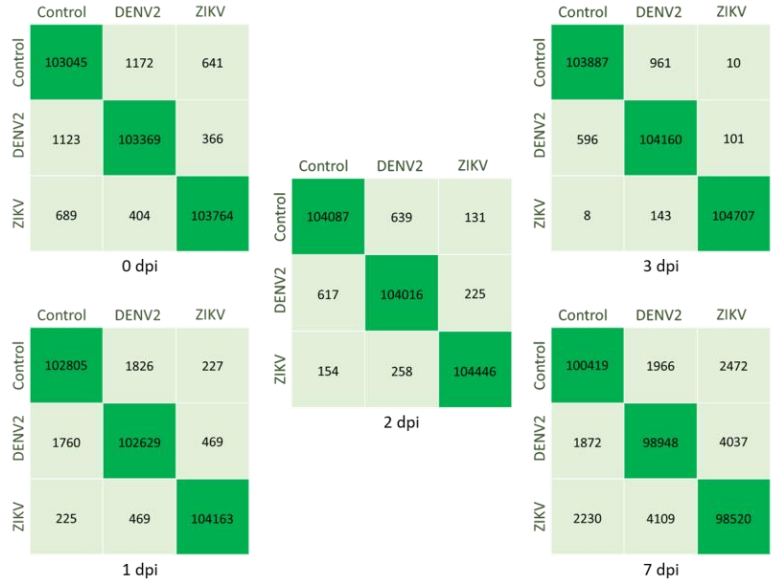

Figure 5. Confusion matrixes for 0, 1, 2, 3, and 7 dpi

The proposed method demonstrated its highest accuracy, precision, recall, and F1 score, reaching 99.42% on the $3^{rd}$ day post-infection. Furthermore, impressive accuracy, precision, recall, and F1 score values exceeding 90% were consistently observed for all other days post-infection. The average accuracy, precision, recall, and F1 score for all days post-infection was 98.1%, while the average training time was 6:13 (Table 6).

Table 6: Evaluation Metrics for the proposed method

| Days Post-Infection | Accuracy (%) | Precision (%) | Recall (%) | F1 Score (%) | Training Time (m:s) |
|---|---|---|---|---|---|
| 0 | 98.60 | 98.60 | 98.60 | 98.60 | 05:56 |
| 1 | 98.42 | 98.42 | 98.42 | 98.42 | 07:46 |
| 2 | 99.36 | 99.36 | 99.36 | 99.36 | 05:18 |
| 3 | 99.42 | 99.42 | 99.42 | 99.42 | 05:29 |
| 7 | 94.70 | 94.70 | 94.70 | 94.70 | 06:37 |
| Average | 98.10 | 98.10 | 98.10 | 98.10 | 6:13 |

The precision-recall curves for 0, 1, 2, 3, and 7 dpi for all 3 classes Control, DENV2, and ZIKV are presented in Figure 6. The precision-recall curves showcase exceptionally good performance, reflecting high precision and recall values across various thresholds.

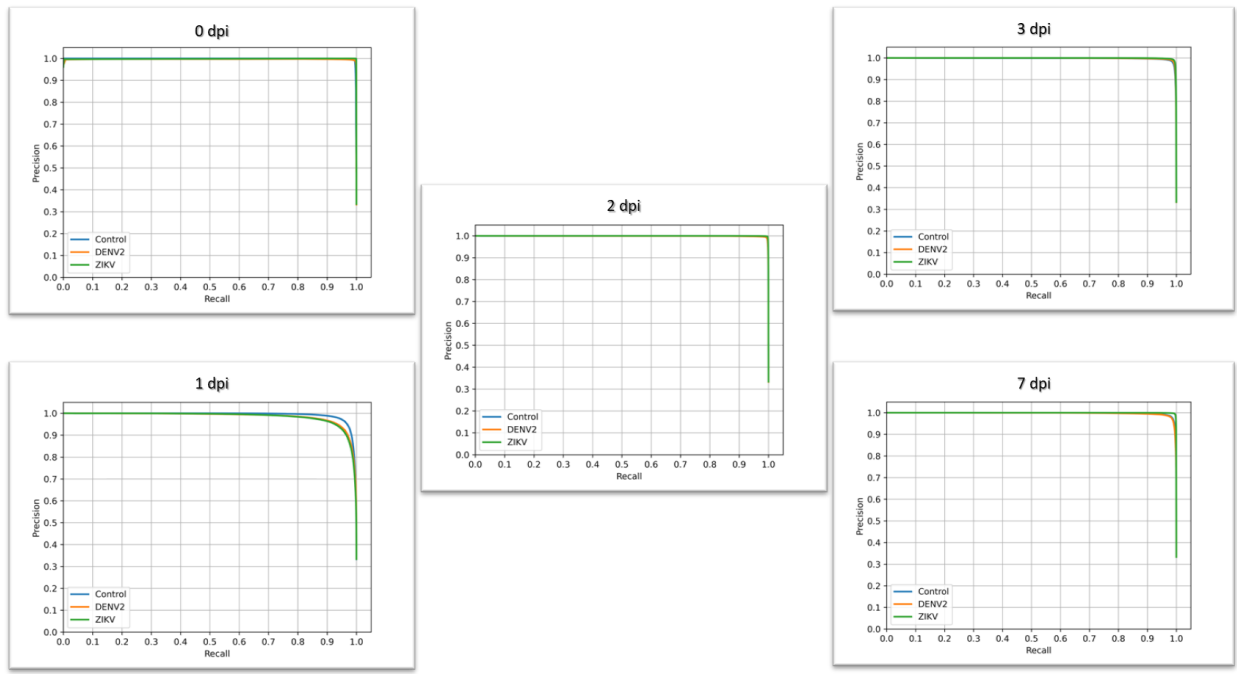

Figure 6: Precision-recall curves for 0, 1, 2, 3, and 7 dpi

## Comparative Analysis

To ensure the capability of the proposed method, the method was also compared with other machine learning methods. The method was compared to 8 different machine learning methods: CNN, Multi-Layer Perceptron (MLP), XGBoost, AdaBoost, Random Forest, Decision Tree, Nive Bayesian and Logistic Regression. All methods were trained using the same dataset. The proposed method exhibited superior performance by consistently surpassing all other methods, achieving accuracy, precision, recall, and F1 scores higher than 94% across all days post-infection. The closest competitor was CNN, with accuracy, precision, recall, and F1 scores higher than 86% across all days post-infection. The naive Bayesian method showed the lowest performance, recording accuracy, precision, recall, and F1 scores below 57% across all days post-infection (Table 7).

Table 7: Evaluation metrics for all machine learning methods.

| Days Post Infection | Evaluation Metrics | CNN | MLP | XGBoost | AdaBoost | Random Forest | Decision Tree | Naive Bayesian | Logistic Regression | **Proposed Method** |
|---|---|---|---|---|---|---|---|---|---|---|
| 0 dpi | Accuracy (%) | 92.68 | 78.54 | 92.64 | 66.50 | 89.21 | 81.31 | 40.98 | 57.40 | **98.60** |
| | Precision (%) | 92.67 | 83.47 | 92.74 | 67.76 | 89.38 | 81.31 | 39.28 | 56.86 | **98.60** |
| | Recall (%) | 92.68 | 78.55 | 92.64 | 66.48 | 89.21 | 81.31 | 40.94 | 57.38 | **98.60** |
| | F1 Score (%) | 92.67 | 78.55 | 92.65 | 66.04 | 89.25 | 81.31 | 37.16 | 57.07 | **98.60** |
| | Training Time (m:s) | 02:35 | 35:14 | 03:21 | 10:43 | 04:29 | 04:25 | 00:01 | 00:29 | **05:56** |
| 1 dpi | Accuracy (%) | 94.44 | 89.29 | 93.76 | 70.68 | 91.20 | 85.82 | 45.74 | 57.01 | **98.42** |
| | Precision (%) | 94.45 | 91.12 | 93.89 | 71.98 | 91.31 | 85.82 | 47.98 | 55.59 | **98.42** |
| | Recall (%) | 94.44 | 89.28 | 93.76 | 70.67 | 91.21 | 85.82 | 45.77 | 57.03 | **98.42** |
| | F1 Score (%) | 94.45 | 89.42 | 93.79 | 70.74 | 91.24 | 85.82 | 43.46 | 56.02 | **98.42** |
| | Training Time (m:s) | 02:49 | 16:15 | 03:13 | 07:12 | 03:51 | 04:26 | 00:01 | 00:29 | **07:46** |
| 2 dpi | Accuracy (%) | 96.53 | 92.34 | 96.07 | 74.45 | 93.72 | 88.75 | 55.41 | 56.11 | **99.36** |
| | Precision (%) | 96.54 | 92.64 | 96.13 | 74.94 | 93.80 | 88.75 | 55.79 | 55.76 | **99.36** |
| | Recall (%) | 96.53 | 92.34 | 96.07 | 74.46 | 93.72 | 88.76 | 55.44 | 56.09 | **99.36** |
| | F1 Score (%) | 96.54 | 92.33 | 96.08 | 74.17 | 93.74 | 88.76 | 54.83 | 55.90 | **99.36** |
| | Training Time (m:s) | 02:43 | 16:02 | 03:10 | 07:12 | 03:34 | 04:23 | 00:01 | 00:30 | **05:18** |
| 3 dpi | Accuracy (%) | 97.21 | 96.45 | 97.68 | 87.09 | 96.75 | 94.54 | 57.57 | 51.30 | **99.42** |
| | Precision (%) | 97.31 | 96.65 | 97.78 | 88.52 | 96.85 | 94.54 | 56.97 | 51.05 | **99.42** |
| | Recall (%) | 97.21 | 96.45 | 97.68 | 87.08 | 96.75 | 94.54 | 57.61 | 51.31 | **99.42** |
| | F1 Score (%) | 97.23 | 96.48 | 97.69 | 87.35 | 96.77 | 94.54 | 55.15 | 51.17 | **99.42** |
| | Training Time (m:s) | 03:51 | 15:07 | 03:20 | 07:16 | 03:22 | 03:50 | 00:01 | 00:27 | **05:29** |
| 7 dpi | Accuracy (%) | 86.13 | 80.73 | 86.70 | 64.17 | 82.24 | 73.39 | 46.94 | 35.45 | **94.70** |
| | Precision (%) | 86.33 | 81.21 | 86.95 | 65.68 | 82.34 | 73.38 | 49.60 | 34.43 | **94.70** |
| | Recall (%) | 86.13 | 80.73 | 86.70 | 64.18 | 82.24 | 73.40 | 46.97 | 35.44 | **94.70** |
| | F1 Score (%) | 86.18 | 80.87 | 86.76 | 63.98 | 82.24 | 73.39 | 44.22 | 31.31 | **94.70** |
| | Training Time (m:s) | 02:46 | 0:14 | 03:13 | 07:02 | 04:29 | 05:08 | 00:01 | 00:09 | **06:37** |

## 5. DISCUSSION

In this section, other state-of-the-art methods are given in Table 8 to have a better sight of the proposed method. It should be considered that this paper introduced a machine learning method on the MEA dataset of insects for the first time. Although other papers were represented in Table 8, none of them utilized this dataset of insects using machine learning techniques. Thus, other research papers were mentioned in the table to have a better understanding of recent trends in the field of spike classification on signal datasets using machine learning algorithms. However, it does not mean the proposed method was implemented exactly in the same way and dataset.

Table 8: Comparing recent trends of machine learning techniques in spike classification.

| Reference | Year | Classification Method | Accuracy (%) | Precision (%) | Recall (%) | F1 Score (%) |
|---|---|---|---|---|---|---|
| [41] | 2020 | Random Forest | 96.9 | 96.5 | 97.4 | - |
| [42] | 2021 | Long Short-Term Memory (LSTM) | - | 85.75 | 92.04 | 88.54 |
| [28] | 2021 | Catboost and Random Forest | 98.00 | 99.50 | 96.78 | - |
| [31] | 2022 | CNN | 99.1 | 98.4 | 99.8 | - |
| [30] | 2023 | CNN | 97.57 | - | 98.90 | - |
| [33] | 2023 | ANN and SNN | 90.05 | 98.60 | 99.40 | - |
| [32] | 2023 | Thresholding-based algorithm | 94.32 | 87.50 | 100.00 | 93.33 |
| **This Work** | **2024** | **Proposed Method** | **98.10** | **98.10** | **98.10** | **98.10** |

Compared to other state-of-the-art methods, it can be observed that the proposed method has a high performance in detecting infection. Among the reasons for the favorable performance of this method, it can be said that the proposed method uses strong preprocessing in its algorithm. Obviously, in machine learning methods, data preprocessing is completely vital. Additionally, the proposed method can be effective in the classification section as well. First, features are extracted by CNN, which is one of the powerful methods of deep learning. Then, the final classification is also performed by the XGBoost method, which is one of the most successful methods of machine learning. According to the mentioned context, the advantages and limitations of the proposed method can be discussed in detail.

Regarding the advantages, in this paper, a new MEA dataset was introduced, and it mentioned how to deal with a dataset with a wide range of spikes for an impressive classification. Although in recent years using some low-pass filtering was trended as a preprocessing phase, we indicated that our recommended technique could be dramatically effective. Moreover, we presented a hybrid method for the classification phase which caused more functionality in comparison with other methods. Thus, the advantages of presenting this research are listed below:

• Introducing a new MEA dataset consisting of more than 15 million samples.
• Presenting a novel preprocessing technique to overcome spiky datasets.
• Nominating a combined procedure method for signal classification.

- Comparing the efficiency of the proposed method with other well-known machine learning algorithms and determining its superiorities.

In terms of limitations, we used a strong type of deep learning algorithm called CNN. Deep learning techniques are quite time-consuming in the training phase. Specifically, when the dataset is huge such as our dataset which contains more than 15 million samples. Fortunately, deep learning techniques can function in real-time in the testing phase and the massive amount of data in training sets can cause more efficiency in the testing phase. The more data in the training set we have, the more time in the training phase we need, and the more functionality in the testing phase it has. Therefore, it is highly recommended that when the proposed method is supposed to be utilized in biological institutions and organizations, the transfer learning technique is used for CNN. That is, after training the CNN model, it should be saved as a pre-trained model. Then, the pre-trained model is used in the proposed method as the CNN in the testing phase. Using this technique, the training phase, which is a completely time-consuming phase will be removed in real industrial usage. Therefore, the limitations of this research are also listed below:

- Being time-consuming of the deep learning' training section for the proposed method.
- Functioning in real-time using transfer learning for industrial applications.

## 6. CONCLUSION

Diseases transmitted via mosquitoes significantly impact public health worldwide, requiring an understanding of the infection dynamics and impact of infection on the vector. Understanding the neural activity of infected mosquitoes is a crucial aspect of unraveling the infection dynamics. Here, we have presented a classification method to classify the neural activity of the Zika-infected, dengue-infected, and uninfected control *Aedes aegypti* mosquitoes. The dataset used in this research comprised more than 15 million samples, encompassing Control, DENV2, and ZIKV classes across various time points, 0, 1, 2, 3, and 7 days post-infection. The method was validated using k-fold cross-validation to ensure its consistency. The method employs a fusion of normalization, feature importance, and dimension reduction for the preprocessing and combines CNN and XGBoost for classification. The performance of the proposed method was evaluated using accuracy, precision, recall, and the F1 scores. The results obtained from the method highlight its remarkable performance, achieving an average accuracy of 98.1%. The performance was also compared with 8 other machine learning algorithms, namely CNN, MLP, XGBoost, AdaBoost, Random Forest, Decision Tree, Naive Bayesian, and Logistic Regression to test the method's capability. The method outperformed the performances of all other machine learning algorithms. This research serves as an efficient method to classify the neural spikes of *Aedes aegypti* mosquitoes, which can assist in understanding the complex interactions between pathogens and mosquitoes. Future work can focus on uncovering the correlation between neural activity and behavioral patterns of mosquitoes.